\documentclass[3p,sort & compress]{elsarticle}
\usepackage[utf8]{inputenc}
\usepackage{amssymb}
\usepackage{latexsym}
\usepackage{amsfonts}
\usepackage{amsthm}
\usepackage{amsbsy}
\usepackage{amsmath}
\usepackage{graphics,graphicx}
\usepackage{subfigure}
\usepackage{multirow,multicol}
\usepackage{bm}
\usepackage{bbm}
\usepackage{color}
\usepackage{float}
\usepackage{calc}
\usepackage{caption}
\usepackage{dsfont}
\usepackage{ifthen}
\usepackage{mathrsfs}
\usepackage[colorlinks,linkcolor=blue,citecolor=blue]{hyperref}
\usepackage{epstopdf}
\usepackage{epsfig}
\usepackage{algorithm}
\usepackage{algorithmic}
\usepackage{pifont}
\usepackage{overpic}
\usepackage{psfrag}
\usepackage{rotating}
\usepackage{stmaryrd}
\usepackage{verbatim}
\usepackage{setspace}
\usepackage{pgf}
\usepackage{tikz}

\newdefinition{remark}{Remark}

\newdefinition{method}{Method}
\newdefinition{example}{Example}

\numberwithin{equation}{section}

\numberwithin{theorem}{section}
\journal{Journal of \LaTeX\ Templates}
\begin{document}
	\begin{frontmatter}
		\title{Multi-scale Receptive Fields Graph Attention Network for Point Cloud Classification}
		\author[shjtmath]{Xi-An Li}\ead{lixa0415@sjtu.edu.cn}
		\author[shjtmath,shjtnature]{Lei Zhang\corref{zl,lj,wly}}\ead{lzhang2012@sjtu.edu.cn}
		\author[FSHospitial]{Li-Yan Wang\corref{zl,lj,wly}}\ead{wangliyankmmc@163.com}
		\author[GIAAET]{Jian Lu}\ead{luj@giaaet.com}
		\cortext[zl,lj,wly]{Corresponding author.}
		\address[shjtmath]{School of Mathematical Sciences, Shanghai Jiao Tong University, Shanghai 200240, PR China}
		\address[shjtnature]{Institute of Natural Sciences, Shanghai Jiao Tong University, Shanghai 200240, PR China}
		\address[FSHospitial]{Department of Stomatology, Foshan Woman and Children’s Hospital, Foshan, Guangdong, 528000, PR China}
		\address[GIAAET]{ Guangdong Institute of Aeronautics and Astronautics Equipment $\&$ Technology, Zhuhai 519000, PR China}
		\begin{abstract}
			Understanding the implication of point cloud is still challenging to achieve the goal of classification or segmentation due to the irregular and sparse structure of point cloud. As we have known, PointNet architecture as a ground-breaking work for point cloud which can learn efficiently shape features directly on unordered 3D point cloud and have achieved favorable performance. However, this model fail to consider the fine-grained semantic information of local structure for point cloud. Afterwards, many valuable works are proposed to enhance the performance of PointNet by means of semantic features of local patch for point cloud. In this paper, a multi-scale receptive fields graph attention network (named after MRFGAT) for point cloud classification is proposed. By focusing on the local fine features of point cloud and applying multi attention modules based on channel affinity, the learned feature map for our network can well capture the abundant features information of point cloud. The proposed MRFGAT architecture is tested on ModelNet10 and ModelNet40 datasets, and results show it achieves state-of-the-art performance in shape classification tasks.
		\end{abstract}
	\end{frontmatter}
	
	\section{Introduction}\label{sec:01}
	Point cloud  as a simple and efficient representation for 3D shapes and scenes which has become more and more popular in the fields of both academia and industry. For example, autonomous vehicle \cite{zhou2018voxelnet,qi2018frustum,ku2017joint, wang2018deep,liang2018deep}, robotic mapping and navigation \cite{biswas2012depth, zhu2017target}, 3D shape representation and modelling \cite{Golovinskiy2009Shape}. Lots of ways can be used to obtain 3D point cloud data, such as utilizing 3D scanners including physical touch or non-contact measurements with light, sound, LiDAR etc.
	
	Up to now, various approaches have been developed to handle this kind of data including traditional handcraft algorithms \cite{mitra2004registration,vosselman20013d,Rusu2009Fast,Tombari2010Unique} and attractive neural networks \cite{charles2017pointnet,pointCNN2018,wang2019dynamic,zhao2019pointweb,thomas2019kpconv}. In terms of these methods, it is significant to classify or segment point cloud by choosing salient features of point cloud, such as normals, curvatures and colors. Handcrafted features are usually employed to address specific problems but difficult to generalize to new tasks. With the development of deep learning, the existed end-to-end neural networks have overcame many challenges stem from 3D data and made great breakthrough for point cloud. In particular, the modificatory works of convolutional neural networks (CNNs) have achieved significant success for point cloud data in computer vision tasks, such as PointNet\cite{charles2017pointnet} and its improved version\cite{qi2017pointnet}, PointCNN\cite{pointCNN2018,atzmon2018point}, PointSift\cite{jiang2018pointsift} and so on. Unfortunately, many neural networks for point cloud only capture global feature without local information which is also an import semantic feature for point cloud. Hence, how to exploit the local informations of point cloud has become a new research hotspot and some valuable works also have been proposed in recent. PointNet++ \cite{qi2017pointnet} extends PointNet model by constructing a hierarchical neural network that recursively applies PointNet with designed sampling and grouping layers to extract local features. Graph neural networks \cite{2005anew,2009graph} can not only directly address a more general class of graphs, e.g. cyclic, directed and undirected graphs, but also be applied to deal with point cloud data. Recent, DGCNN \cite{wang2019dynamic} and its variant \cite{zhang2019linked} well utilized the graph networks with respect to the edges convolution on points, then obtained the local edges information of point cloud. Other relevant works applying graph structure of point cloud can be found in \cite{te2018rgcnn,gao2019exploring,lu2020pointngcnn}.
	
	\begin{figure}
		\centering
		\includegraphics[scale=0.65]{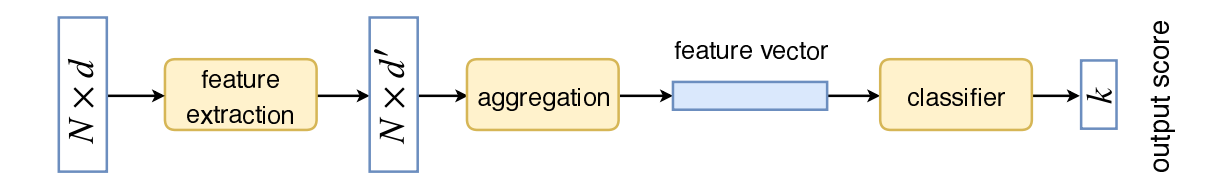}
		\caption{A typical architecture of a 3D point cloud classification framework. The model takes $N$ points of dimension $d$ as its input, then extracts point features of dimension $d'$, followed by an aggregation module used to build a feature vector which is invariant to point permutation. Lastly, a classifier is used to classify the resulting feature vector into one of the $k$ classes.}
	\end{figure}
	
	Attention mechanism plays a significant role in machine translation task \cite{vaswani2017attention}, vision-based task \cite{mnih2014recurrent}, and graph-based task \cite{GAT2017graph}. Combining graph structure and attention mechanism, some favorable networks architectures are constructed which leverage well the local semantic feature of point cloud. Readers can refer to \cite{chen2019gapnet,wang2019graph,feng2019point,kang2019pyramnet}.
	
	Inspired by graph attention networks\cite{GAT2017graph}, graph convolution network \cite{niepert2016learning} and local contextual information networks such as DGCNN\cite{wang2019dynamic,zhang2019linked} and GAPNet\cite{chen2019gapnet}, we design a multi-scale receptive fields graph attention networks for point cloud classification. Unlike previous models that only consider the attribute information such as coordinate of each single point or only exploit local semantic information of point, we pay attentions on the spatial context information of both local and global structure for point cloud. Finally, like the standard convolution in grid domain, our model can also be efficiently implemented for the graph representation of a point cloud.
	
	The key contributions of our work are summarized as follows:
	\begin{itemize}
		\item We construct graph of local patch for point cloud, then enhance the feature representation of point in point cloud by combining edges' information and neighbors' information.
		\item We introduce a multi-scale receptive fields mechanism  to capture the local semantic features in various ranges for point cloud.
		\item We balance the influence between neighbors and centroid in local graph by means of attention mechanisms.
		\item We release our code to facilitate reproducibility and future research.
	\end{itemize}
	
	The rest parts of this paper are structured as follows. In section \ref{sec:02}, we review the most closely related literatures on point cloud. Then in Section \ref{sec:03} we present our proposed MRFGAT architecture and provide the details of the networks in terms of shape classification for point cloud. We describe the dataset and design comparison algorithms in Section \ref{sec:04}, followed by the experiments results and discussion. Finally, some concluding remarks are made in Section \ref{sec:05}.
	
	\section{RELATED WORKS}\label{sec:02}
	\subsection{Pointwise MLP and Point convolution networks}\label{sec:02:1}
	Utilizing the deep learning technique, the classical PointNet \cite{charles2017pointnet} was proposed to deal with directly unordered point clouds without using any volumetric or grid-mesh representation. The main idea of this network are as follows. At first, a Spatial Transformer Networks (STN) module similar to feature-extracting process is constructed which guarantees the invariance of transformations; Secondly, a shared pointwise Multi-Layer-Perceptron(MLP) module  is introduced which is used to extract semantic features form point sets; At last, the final semantic information of point cloud are aggregated by means of a max pooling layer. Due to the favorable ability to approximate any continuous function for MLP which is easy to implement by point convolution, some related works were presented according to the PointNet architecture \cite{li2018so-net,yu2018pu}.
	
	Similar to convolution operator in 2D space, some convolution kernels for points in 3D space are designed which can capture the abundant information of point cloud. PointCNN\cite{pointCNN2018} used a local $\mathcal{X}$--transformation kernel to fulfill the invariance of permutation for points, then generalize this technique to hierarchical form in analogy to that of image CNNs. \cite{klokov2017escape,you2018pointwise,wu2019pointconv} extended the convolution operator of 2D space and applied at individual point in local region of point cloud, then collected the neighbors' information in hierarchical convolution layer to the center point. Kernel Point Convolution (KPConv) \cite{thomas2019kpconv} consists of a set of local 3D filters and overcomes stand point convolution limitation. This novel kernel structure is very flexible to learn local geometric patterns without any weights.
	
	\subsection{Learning local features}\label{sec:02:2}
	In order to overcome the shortcoming of failing to use local features for PointNet-like networks, some hierarchical architectures have been developed to aggregate local information with MLP by considering local spatial relationships of 3D data, such as \cite{qi2017pointnet,li2018so-net}. In contrast to the previous type, these method can avoid sparsity and update dynamically in different feature dimensions. According to a Capsule Networks, 3D Capsule Convolutional Networks were developed which can learn well the local features of point cloud, one can refer to \cite{zhao20193d,srivastava2019geometric,cheraghian20193dcapsule}.
	
	\subsection{Graph Convolutional Networks }\label{sec:02:4}
	Graph Convolutional Neural Networks (GCNNs) have become more and more attraction to address irregularly structured data, such as citation networks and social networks. In terms of 3D point clouds data, GCNNs have shown its powerful ability on classification and segmentation. Using the convolution with respect to graph in the spectral domain is an important approach \cite{xu2018spidercnn,boscaini2015learning,yi2017syncspeccnn}. But, it needs to calculate a lot of parameters on polynomial or rational spectral filters \cite{defferrard2016convolutional}. Recent, many researchers constructed  local graph by applying each point's neighbors in embedding space based on $N$-dimensional Euclidean distance, then grouped each point's neighbors in the form of high dimensional vectors, such as EdgeConv-like works \cite{wang2019dynamic,zhang2019linked,dynamicScale2019} and graph convolutions \cite{niepert2016learning,verma2018feastnet}. Compared with the spectral methods, its main merit is that it is more consistent with the characteristics of data distribution. Specially, EdgeConv extracts edge features through the relationship between  central point and  neighbor points by successively constructing graph in hierarchical model. To sum up, the graph convolution networks combine features on local surface patches which are invariant to the deformations of patches for point cloud in Euclidean space.

	\subsection{Attention mechanism}\label{sec:02:3}
	The idea of attention has been successfully used in natural language processing(NLP) \cite{vaswani2017attention} and graph-based work \cite{GAT2017graph,lee2018attention}, etc.. Attention module can balance the weight relationship of different nodes in graph structure data or different parts in sequence data.
	
	Recent, the attention idea  has obtained more  and more attraction and made a great contribution to point clouds learning works \cite{chen2019gapnet,wang2019graph}. In these works, it is significant to aggregate point or edge features by means of attention module. Differently from existing methods, we try to enhance the high-level representation of point cloud by capturing the relation of points and local fine-information along its channels.
	
	\section{Our approach}\label{sec:03}
	The classification of point cloud includes two contents: taking the 3D point cloud as input and assigning one semantic class label for each point. Based on the technique of extracting features from local directed graph and attention mechanism, a new architecture is proposed to better learn point's representation for unstructured point cloud in shape classification task. This new model consists of three components which are the point enrichment, the feature representation and the prediction. These three components fully couple together, ensuring an end-to-end training manner.
	
	\subsection{Problem statement}
	At first, we let $P = \{p_i\in \mathbb{R}^F,i=1,2,\cdots,N\}$ represent a raw set of unordered points which as the input for our mode, where $N$ is the number of the points and $p_i$ is a feature vector with a dimension $F$. In actual applications, the feature vector $p_i$ might contain 3D space coordinates $(x,y,z)$, color, intensity, surface normal, etc. For the sake of simplicity, we set $F = 3$ in our work and only choose 3D coordinates of point as the point feature. A classification or semantic segmentation of a point cloud are function $\Phi_c$ or $\Phi_s$ which assign individual point semantic labels or point cloud semantic labels, respectively, i.e.,
	\begin{equation*}
	\Phi:P\rightarrow L^k
	\end{equation*}
	Here, $\Phi$ represents $\Phi_c$ or $\Phi_s$. The objective of algorithms are finding optimal function that gives accurate semantic labels.
	
	There are several design constraints for the classification function $\Phi_c$ and segmentation function $\Phi_s$. 1) Permutation invariance: the order of points may
	vary but does not influence the category of the point cloud; 2)Transformation invariance: the results of classification or segmentation should not be changed owing to the translation and rotation of generated point cloud.
	
	\subsection{Graph generation for Point Cloud}
		\begin{figure}[!htp]
		\centering
		\begin{tikzpicture}	[scale=0.75]
		\node[circle, fill=red!100] (pi) at (0,0){$p_i$};
		\node[circle, fill=blue!50] (pi2) at (2,1){$p_{i2}$};
		\node[circle, fill=blue!50] (pi6) at (-2,1){$p_{i6}$};
		\node[circle, fill=blue!50] (pi3) at (2,-1){$p_{i3}$};
		\node[circle, fill=blue!50] (pi5) at (-2,-1){$p_{i5}$};
		\node[circle, fill=blue!50] (pi1) at (0,2.5){$p_{i1}$};
		\node[circle, fill=blue!50] (pi4) at (0,-2.5){$p_{i4}$};
		
		\draw[line width=0.9pt,color=cyan,-] (pi1) -- (pi);
		\draw[line width=0.9pt,color=cyan,-] (pi2) -- (pi);
		\draw[line width=0.9pt,color=cyan,-] (pi3) -- (pi);
		\draw[line width=0.9pt,color=cyan,-] (pi4) -- (pi);
		\draw[line width=0.9pt,color=cyan,-] (pi5) -- (pi);
		\draw[line width=0.9pt,color=cyan,-] (pi6) -- (pi);
		
		\draw[line width=0.9pt,color=cyan,-] (pi1) -- (pi2);
		\draw[line width=0.9pt,color=cyan,-] (pi1) -- (pi3);
		\draw[line width=0.9pt,color=cyan,-] (pi1) -- (pi5);
		\draw[line width=0.9pt,color=cyan,-] (pi1) -- (pi6);
		
		\draw[line width=0.9pt,color=cyan,-] (pi2) -- (pi3);
		\draw[line width=0.9pt,color=cyan,-] (pi2) -- (pi4);
		\draw[line width=0.9pt,color=cyan,-] (pi2) -- (pi6);
		
		\draw[line width=0.9pt,color=cyan,-] (pi3) -- (pi4);
		\draw[line width=0.9pt,color=cyan,-] (pi3) -- (pi5);
		
		\draw[line width=0.9pt,color=cyan,-] (pi4) -- (pi5);
		\draw[line width=0.9pt,color=cyan,-] (pi4) -- (pi6);
		
		\draw[line width=0.9pt,color=cyan,-] (pi5) -- (pi6);
		\end{tikzpicture}
		$~~~~~~~~~$
		\begin{tikzpicture}	[scale=0.75]
		\node[circle, fill=red!100] (pi) at (0,0){$p_i$};
		\node[circle, fill=blue!50] (pi2) at (2,1){$p_{i2}$};
		\node[circle, fill=blue!50] (pi6) at (-2,1){$p_{i6}$};
		\node[circle, fill=blue!50] (pi3) at (2,-1){$p_{i3}$};
		\node[circle, fill=blue!50] (pi5) at (-2,-1){$p_{i5}$};
		\node[circle, fill=blue!50] (pi1) at (0,2.5){$p_{i1}$};
		\node[circle, fill=blue!50] (pi4) at (0,-2.5){$p_{i4}$};
		\draw[line width=0.9pt,color=magenta,->] (pi1) --node[left]{$e_{i1}$} (pi);
		\draw[line width=0.9pt,color=magenta,->] (pi2) --node[above]{$e_{i2}$} (pi);
		\draw[line width=0.9pt,color=magenta,->] (pi3) --node[below]{$e_{i3}$} (pi);
		\draw[line width=0.9pt,color=magenta,->] (pi4) --node[left]{$e_{i4}$} (pi);
		\draw[line width=0.9pt,color=magenta,->] (pi5) --node[below]{$e_{i5}$} (pi);
		\draw[line width=0.9pt,color=magenta,->] (pi6) --node[above]{$e_{i6}$} (pi);
		\end{tikzpicture}
		\caption{Graph of point cloud. The $p_i$ and $\{p_{i1},p_{i2},p_{i3}, p_{i4},p_{i5},p_{i6}\}$ are a central point and its neighbors, respectively. The directed edges from the neighbors to the central point are denoted by $\{e_{i1},e_{i2},e_{i3}, e_{i4}, e_{i5},e_{i6}\}$.}
		\label{fig:Lshape}
	\end{figure}
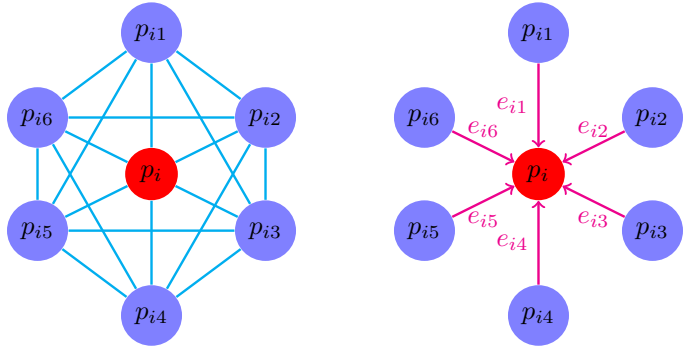

	Some works indicate that local features of point cloud can help to improve the discriminability of point, then exploring the relationship among points in a local region is the keypoint for this paper. Graph neural network is a feasible approach to process point cloud, because it propagates on each node individually and ignores the input order of nodes, then extracts the local information of dependency between nodes. To apply graph neural network on the point cloud, we need to convert it to a directed graph. Like DGCNN \cite{wang2019dynamic,zhang2019linked} and GAPNet \cite{chen2019gapnet}, we obtain the neighbors(including self) of each point in point cloud by means of $K$-NN algorithm before convolutional operation, then construct a local directed graph in Euclidean space. In the directed graph $G = (V; E)$ of local patch for point cloud, $V = \{1,2,\cdots,K\}$ are the vertices of $G$, namely, the nodes of point cloud, $E$ stands for the edge set of $G$ and each edge is $e_{ij}=p_i-p_{ij}$ with $p_i\in P$ and $p_{ij}\in V$ being centroid and neighbors, respectively.

	\subsection{Single Receptive Field Graph Attention Layer(SRFGAT)}		
	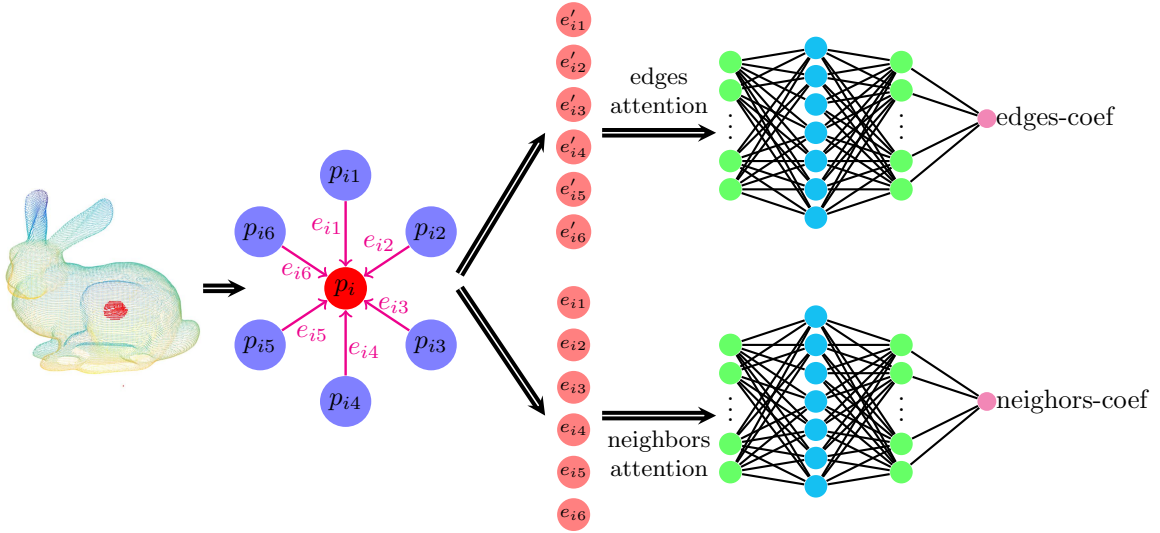
\begin{figure}[H]
		\begin{center}
			\usetikzlibrary{decorations.text}
			\begin{tikzpicture}[scale=0.75]	
			\node at (-5,0){\includegraphics[scale=0.125]{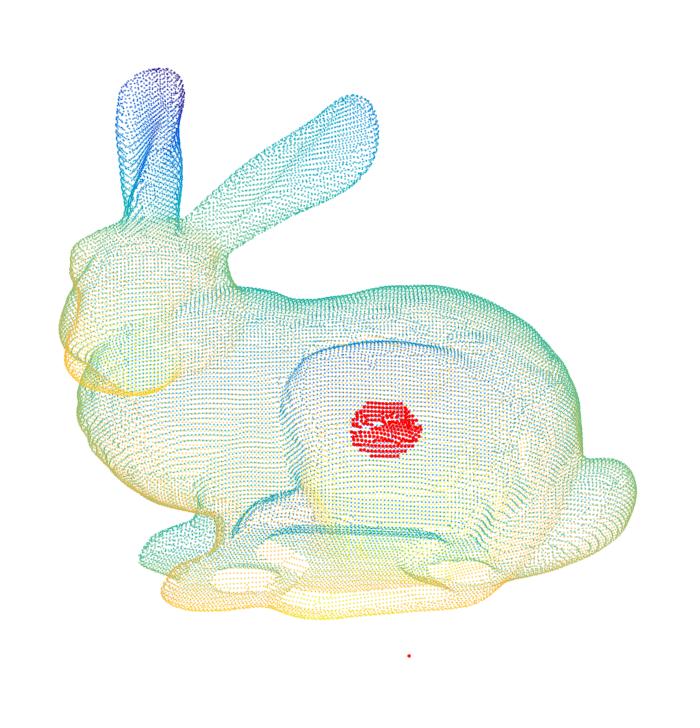}};
			\coordinate (tuzi1) at (-3.25, 0.0);			
			\coordinate (tuzi2) at (-2.5, 0.0);
			\path[-stealth, line width=1.5pt] (tuzi1) edge [double] (tuzi2);	
			
			\node[circle, fill=red!100,inner sep=2pt] (pi) at (-0.75,0){$p_i$};
			\node[circle, fill=blue!50,inner sep=2pt] (pi2) at (0.75,1){$p_{i2}$};
			\node[circle, fill=blue!50,inner sep=2pt] (pi6) at (-2.25,1){$p_{i6}$};
			\node[circle, fill=blue!50,inner sep=2pt] (pi3) at (0.75,-1){$p_{i3}$};
			\node[circle, fill=blue!50,inner sep=2pt] (pi5) at (-2.25,-1){$p_{i5}$};
			\node[circle, fill=blue!50,inner sep=2pt] (pi1) at (-0.75,2.0){$p_{i1}$};
			\node[circle, fill=blue!50,inner sep=2pt] (pi4) at (-0.75,-2.0){$p_{i4}$};
			
			\draw[line width=0.9pt,color=magenta,->] (pi1) --node[left, xshift=1mm, yshift=1mm]{$e_{i1}$} (pi);
			\draw[line width=0.9pt,color=magenta,->] (pi2) --node[above, xshift=-1mm]{$e_{i2}$} (pi);
			\draw[line width=0.9pt,color=magenta,->] (pi3) --node[above, xshift=1mm, yshift=-1mm]{$e_{i3}$} (pi);
			\draw[line width=0.9pt,color=magenta,->] (pi4) --node[right, xshift=-1mm, yshift=-1mm]{$e_{i4}$} (pi);
			\draw[line width=0.9pt,color=magenta,->] (pi5) --node[below, xshift=1mm]{$e_{i5}$} (pi);
			\draw[line width=0.9pt,color=magenta,->] (pi6) --node[below, xshift=-1mm, yshift=1mm]{$e_{i6}$} (pi);
			
			\node[circle, fill=red!50,inner sep=0.1pt] (new-edge1) at (3.25,4.75){\scriptsize$e'_{i1}$};
			\node[circle, fill=red!50,inner sep=0.1pt] (new-edge2) at (3.25,4.0){\scriptsize$e'_{i2}$};
			\node[circle, fill=red!50,inner sep=0.1pt] (new-edge3) at (3.25,3.25){\scriptsize$e'_{i3}$};
			\node[circle, fill=red!50,inner sep=0.1pt] (new-edge4) at (3.25,2.5){\scriptsize$e'_{i4}$};
			\node[circle, fill=red!50,inner sep=0.1pt] (new-edge5) at (3.25,1.75){\scriptsize$e'_{i5}$};
			\node[circle, fill=red!50,inner sep=0.1pt] (new-edge6) at (3.25,1.0){\scriptsize$e'_{i6}$};
			
			\node[circle, fill=red!50,inner sep=0.5pt] (edge1) at (3.25,-0.25){\scriptsize$e_{i1}$};
			\node[circle, fill=red!50,inner sep=0.5pt] (edge2) at (3.25,-1.0){\scriptsize$e_{i2}$};
			\node[circle, fill=red!50,inner sep=0.5pt] (edge3) at (3.25,-1.75){\scriptsize$e_{i3}$};
			\node[circle, fill=red!50,inner sep=0.5pt] (edge4) at (3.25,-2.5){\scriptsize$e_{i4}$};
			\node[circle, fill=red!50,inner sep=0.5pt] (edge5) at (3.25,-3.25){\scriptsize$e_{i5}$};
			\node[circle, fill=red!50,inner sep=0.5pt] (edge6) at (3.25,-4.0){\scriptsize$e_{i6}$};
			
			\coordinate (point0) at (1.25, 0.25);		
			\coordinate (point1) at (2.75, 2.75);
			\coordinate (point2) at (1.25, 0.0);		
			\coordinate (point3) at (2.75, -2.25);						
			\coordinate (resnet1) at (3.75, 2.75);			
			\coordinate (resnet2) at (5.75, 2.75);			
			\coordinate (resnet3) at (3.75, -2.25);			
			\coordinate (resnet4) at (5.75, -2.25);
			
			\path[-stealth, line width=1.5pt] (point0) edge [double] (point1);	
			\path[-stealth, line width=1.5pt] (point2) edge [double] (point3);		
			\path[-stealth, line width=1.5pt] (resnet1) edge [double]  node [above, rotate=0] {\small $\begin{matrix} \text{edges}\\ \text{attention} \end{matrix}$} (resnet2);			
			\path[-stealth, line width=1.5pt] (resnet3) edge [double]  node [below, rotate=0] {\small $\begin{matrix} \text{neighbors}\\ \text{attention} \end{matrix}$} (resnet4);
			
			\node[circle, fill=green!60,inner sep=3pt] (x0) at (6, 4) {};
			\node[circle, fill=green!60,inner sep=3pt] (x1) at (6, 3.5) {};
			\node[circle, inner sep=1pt] (x2) at (6, 3) {$\vdots$};	
			\node[circle, fill=green!60,inner sep=3pt] (x3) at (6, 2.25) {};
			\node[circle, fill=green!60,inner sep=3pt] (x4) at (6, 1.75) {};

			\node[circle, fill=cyan!70,inner sep=3pt] (h10) at (7.5, 4.25) {};
			\node[circle, fill=cyan!70,inner sep=3pt] (h11) at (7.5, 3.75) {};
			\node[circle, fill=cyan!70,inner sep=3pt] (h12) at (7.5, 3.25) {};
			\node[circle, fill=cyan!70,inner sep=3pt] (h13) at (7.5, 2.75) {};
			\node[circle, fill=cyan!70,inner sep=3pt] (h14) at (7.5, 2.25) {};
			\node[circle, fill=cyan!70,inner sep=3pt] (h15) at (7.5, 1.75) {};
			\node[circle, fill=cyan!70,inner sep=3pt] (h16) at (7.5, 1.25) {};
			
			\draw[line width=0.8pt,-] (x0) --  (h10);
			\draw[line width=0.8pt,-] (x0) --  (h11);
			\draw[line width=0.8pt,-] (x0) --  (h12);
			\draw[line width=0.8pt,-] (x0) --  (h13);
			\draw[line width=0.8pt,-] (x0) --  (h14);
			\draw[line width=0.8pt,-] (x0) --  (h15);
			\draw[line width=0.8pt,-] (x0) --  (h16);
			\draw[line width=0.8pt,-] (x1) --  (h10);
			\draw[line width=0.8pt,-] (x1) --  (h11);
			\draw[line width=0.8pt,-] (x1) --  (h13);
			\draw[line width=0.8pt,-] (x1) --  (h14);
			\draw[line width=0.8pt,-] (x1) --  (h15);
			\draw[line width=0.8pt,-] (x1) --  (h16);
			\draw[line width=0.8pt,-] (x3) --  (h10);
			\draw[line width=0.8pt,-] (x3) --  (h11);
			\draw[line width=0.8pt,-] (x3) --  (h13);
			\draw[line width=0.8pt,-] (x3) --  (h14);
			\draw[line width=0.8pt,-] (x3) --  (h15);
			\draw[line width=0.8pt,-] (x3) --  (h16);
			\draw[line width=0.8pt,-] (x4) --  (h10);
			\draw[line width=0.8pt,-] (x4) --  (h11);
			\draw[line width=0.8pt,-] (x4) --  (h13);
			\draw[line width=0.8pt,-] (x4) --  (h14);
			\draw[line width=0.8pt,-] (x4) --  (h15);
			\draw[line width=0.8pt,-] (x4) --  (h16);
			
			\node[circle, fill=green!60,inner sep=3pt] (y0) at (9, 4) {};
			\node[circle, fill=green!60,inner sep=3pt] (y1) at (9, 3.5) {};
			\node[circle, inner sep=1pt] (y2) at (9, 3) {$\vdots$};	
			\node[circle, fill=green!60,inner sep=3pt] (y3) at (9, 2.25) {};
			\node[circle, fill=green!60,inner sep=3pt] (y4) at (9, 1.75) {};
			
			\draw[line width=0.8pt,-] (h10) --  (y0);
			\draw[line width=0.8pt,-] (h10) --  (y1);
			\draw[line width=0.8pt,-] (h10) --  (y3);
			\draw[line width=0.8pt,-] (h10) --  (y4);
			\draw[line width=0.8pt,-] (h11) --  (y0);
			\draw[line width=0.8pt,-] (h11) --  (y1);
			\draw[line width=0.8pt,-] (h11) --  (y3);
			\draw[line width=0.8pt,-] (h11) --  (y4);
			\draw[line width=0.8pt,-] (h12) --  (y0);
			\draw[line width=0.8pt,-] (h12) --  (y1);
			\draw[line width=0.8pt,-] (h12) --  (y3);
			\draw[line width=0.8pt,-] (h12) --  (y4);
			\draw[line width=0.8pt,-] (h13) --  (y0);
			\draw[line width=0.8pt,-] (h13) --  (y1);
			\draw[line width=0.8pt,-] (h13) --  (y3);
			\draw[line width=0.8pt,-] (h13) --  (y4);
			\draw[line width=0.8pt,-] (h14) --  (y0);
			\draw[line width=0.8pt,-] (h14) --  (y1);
			\draw[line width=0.8pt,-] (h14) --  (y3);
			\draw[line width=0.8pt,-] (h14) --  (y4);
			\draw[line width=0.8pt,-] (h15) --  (y0);
			\draw[line width=0.8pt,-] (h15) --  (y1);
			\draw[line width=0.8pt,-] (h15) --  (y3);
			\draw[line width=0.8pt,-] (h15) --  (y4);
			\draw[line width=0.8pt,-] (h16) --  (y0);
			\draw[line width=0.8pt,-] (h16) --  (y1);
			\draw[line width=0.8pt,-] (h16) --  (y3);
			\draw[line width=0.8pt,-] (h16) --  (y4);
			
			\node[circle,fill=magenta!60, inner sep=2.5pt] (edgecoef) at (10.5, 3) {};
			\node at (10.5,3)[right]{edges-coef};
			
			\draw[line width=0.8pt,-] (y0) --  (edgecoef);
			\draw[line width=0.8pt,-] (y1) --  (edgecoef);
			\draw[line width=0.8pt,-] (y3) --  (edgecoef);
			\draw[line width=0.8pt,-] (y4) --  (edgecoef);	
			
			\node[circle, fill=green!60,inner sep=3pt] (x00) at (6, -1) {};
			\node[circle, fill=green!60,inner sep=3pt] (x10) at (6, -1.5) {};
			\node[circle, inner sep=1pt] (x20) at (6, -2) {$\vdots$};	
			\node[circle, fill=green!60,inner sep=3pt] (x30) at (6, -2.75) {};
			\node[circle, fill=green!60,inner sep=3pt] (x40) at (6, -3.25) {};

			\node[circle, fill=cyan!70,inner sep=3pt] (h20) at (7.5, -0.5) {};
			\node[circle, fill=cyan!70,inner sep=3pt] (h21) at (7.5, -1) {};
			\node[circle, fill=cyan!70,inner sep=3pt] (h22) at (7.5, -1.5) {};
			\node[circle, fill=cyan!70,inner sep=3pt] (h23) at (7.5, -2) {};
			\node[circle, fill=cyan!70,inner sep=3pt] (h24) at (7.5, -2.5) {};
			\node[circle, fill=cyan!70,inner sep=3pt] (h25) at (7.5, -3) {};
			\node[circle, fill=cyan!70,inner sep=3pt] (h26) at (7.5, -3.5) {};
			
			\draw[line width=0.8pt,-] (x00) --  (h20);
			\draw[line width=0.8pt,-] (x00) --  (h21);
			\draw[line width=0.8pt,-] (x00) --  (h22);
			\draw[line width=0.8pt,-] (x00) --  (h23);
			\draw[line width=0.8pt,-] (x00) --  (h24);
			\draw[line width=0.8pt,-] (x00) --  (h25);
			\draw[line width=0.8pt,-] (x00) --  (h26);
			\draw[line width=0.8pt,-] (x10) --  (h20);
			\draw[line width=0.8pt,-] (x10) --  (h21);
			\draw[line width=0.8pt,-] (x10) --  (h23);
			\draw[line width=0.8pt,-] (x10) --  (h24);
			\draw[line width=0.8pt,-] (x10) --  (h25);
			\draw[line width=0.8pt,-] (x10) --  (h26);
			\draw[line width=0.8pt,-] (x30) --  (h20);
			\draw[line width=0.8pt,-] (x30) --  (h21);
			\draw[line width=0.8pt,-] (x30) --  (h23);
			\draw[line width=0.8pt,-] (x30) --  (h24);
			\draw[line width=0.8pt,-] (x30) --  (h25);
			\draw[line width=0.8pt,-] (x30) --  (h26);
			\draw[line width=0.8pt,-] (x40) --  (h20);
			\draw[line width=0.8pt,-] (x40) --  (h21);
			\draw[line width=0.8pt,-] (x40) --  (h23);
			\draw[line width=0.8pt,-] (x40) --  (h24);
			\draw[line width=0.8pt,-] (x40) --  (h25);
			\draw[line width=0.8pt,-] (x40) --  (h26);
			
			\node[circle, fill=green!60,inner sep=3pt] (y00) at (9, -1) {};
			\node[circle, fill=green!60,inner sep=3pt] (y10) at (9, -1.5) {};
			\node[circle, inner sep=1pt] (y20) at (9, -2) {$\vdots$};	
			\node[circle, fill=green!60,inner sep=3pt] (y30) at (9, -2.75) {};
			\node[circle, fill=green!60,inner sep=3pt] (y40) at (9, -3.25) {};
			
			\draw[line width=0.8pt,-] (h20) --  (y00);
			\draw[line width=0.8pt,-] (h20) --  (y10);
			\draw[line width=0.8pt,-] (h20) --  (y30);
			\draw[line width=0.8pt,-] (h20) --  (y40);
			\draw[line width=0.8pt,-] (h21) --  (y00);
			\draw[line width=0.8pt,-] (h21) --  (y10);
			\draw[line width=0.8pt,-] (h21) --  (y30);
			\draw[line width=0.8pt,-] (h21) --  (y40);
			\draw[line width=0.8pt,-] (h22) --  (y00);
			\draw[line width=0.8pt,-] (h22) --  (y10);
			\draw[line width=0.8pt,-] (h22) --  (y30);
			\draw[line width=0.8pt,-] (h22) --  (y40);
			\draw[line width=0.8pt,-] (h23) --  (y00);
			\draw[line width=0.8pt,-] (h23) --  (y10);
			\draw[line width=0.8pt,-] (h23) --  (y30);
			\draw[line width=0.8pt,-] (h23) --  (y40);
			\draw[line width=0.8pt,-] (h24) --  (y00);
			\draw[line width=0.8pt,-] (h24) --  (y10);
			\draw[line width=0.8pt,-] (h24) --  (y30);
			\draw[line width=0.8pt,-] (h24) --  (y40);
			\draw[line width=0.8pt,-] (h25) --  (y00);
			\draw[line width=0.8pt,-] (h25) --  (y10);
			\draw[line width=0.8pt,-] (h25) --  (y30);
			\draw[line width=0.8pt,-] (h25) --  (y40);
			\draw[line width=0.8pt,-] (h26) --  (y00);
			\draw[line width=0.8pt,-] (h26) --  (y10);
			\draw[line width=0.8pt,-] (h26) --  (y30);
			\draw[line width=0.8pt,-] (h26) --  (y40);
			
			\node[circle,fill=magenta!60, inner sep=2.5pt] (neighcoef) at (10.5, -2.0) {};
			\node at (10.5,-2.0)[right]{neighors-coef};
			\draw[line width=0.8pt,-] (y00) --  (neighcoef);
			\draw[line width=0.8pt,-] (y10) --  (neighcoef);
			\draw[line width=0.8pt,-] (y30) --  (neighcoef);
			\draw[line width=0.8pt,-] (y40) --  (neighcoef);
			\end{tikzpicture}
		\end{center}
		\caption{An illustration of attention coefficients generation. The edge feature not only server as local information to center point, but also response the effect of neighbors and centroid. Edge-attention and neighbor-attention reflect the important of edge features and neighbor features to centroid, respectively.}
	\end{figure}
	In order to aggregate the information of neighbors, we use a neighboring-attention mechanism which is introduced to obtain attention coefficients of neighbors for each point. Additionally, edge features are important local features which can enhance the semantic expression of point, then a edge-attention mechanism is also introduced to aggregate information of different edges. In light of the attention mechanism\cite{GAT2017graph,chen2019gapnet}, we firstly transform the neighbors and edges into a high-level feature space to obtain sufficient expressive power. To this end, as an initial step, a parametric non-linear function $h(\cdot)$ is applied to every neighbor and edge, the results are defined by Equation \eqref{eq31} and \eqref{eq32}
	\begin{equation}\label{eq31}
	e'_{ij}=h(e_{ij},\theta)\in \mathbb{R}^{F'}
	\end{equation}
	and
	\begin{equation}\label{eq32}
	p'_{ij}=h(p_{ij},\theta) \in \mathbb{R}^{F'}
	\end{equation}
	respectively, where $\theta$ is a set of learnable parameters of the filter and $F'$ is output dimension. In our method, the function $h(\cdot)$ is set to a single-layer neural network.
	
	It is worthwhile to noting that edges in Euclidean space not only stand for the local features, but also indicate the dependency between centroid and neighbor. We then obtain attentional coefficients of edges and neighbors  by Equation \eqref{eq33}
	\begin{equation}\label{eq33}
	a_{ij}=LeaklyReLU(g(e'_{ij},\theta))~\textup{and}~b_{ij}=LeaklyReLU(g(e_{ij},\theta))
	\end{equation}
	respectively, where $g(e'_{ij},\theta)$ and $g(e_{ij},\theta)$ are single-layer neural network with 1-dimension output. $LeakyReLU(\cdot)$ denotes non-linear activation function leaky ReLU. To make coefficients easily comparable across different neighbors and edges, we use softmax function to normalize the above coefficients which are defined as
	\begin{equation}\label{eq34}
	\alpha_{ij} = \frac{exp(a_{ij})}{\sum_k exp(a_{ik})}~\textup{and}~\beta_{ij} = \frac{exp(b_{ij})}{\sum_k exp(b_{ik})},
	\end{equation}
	respectively, then we use the normalized coefficients to compute contextual feature for every point and it is
	\begin{equation}
	\tilde{x}_i = f\left(\left(\sum_{j}\alpha_{ij}e'_{ij}\right)\bigg{\|}\left(\sum_{j}\beta_{ij}q'_{ij}\right)\right)
	\end{equation}
	where $f(\cdot)$ is a non-linear activation function and $||$ is concatenation operation. In our model, we chose ReLU as $f(\cdot)$.
	
	\subsection{Multi-scale Receptive Fields Graph Attention Layer(MRFGAT).}
	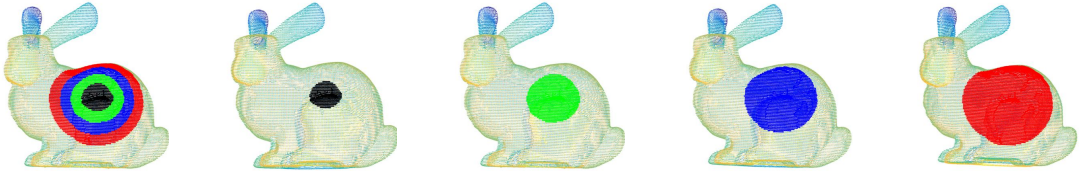
\begin{figure}[H]
		\begin{center}
			\begin{tikzpicture}[scale=0.75]	
			\node (point0) at (-5,0.8){};
			\node (point1) at (-1,0.5){};
			\node (point2) at (3,0.5){};
			\node (point3) at (7,0.5){};
			\node (point4) at (11,0.5){};
			\node (tuzi) at (-5,0){\includegraphics[scale=0.15]{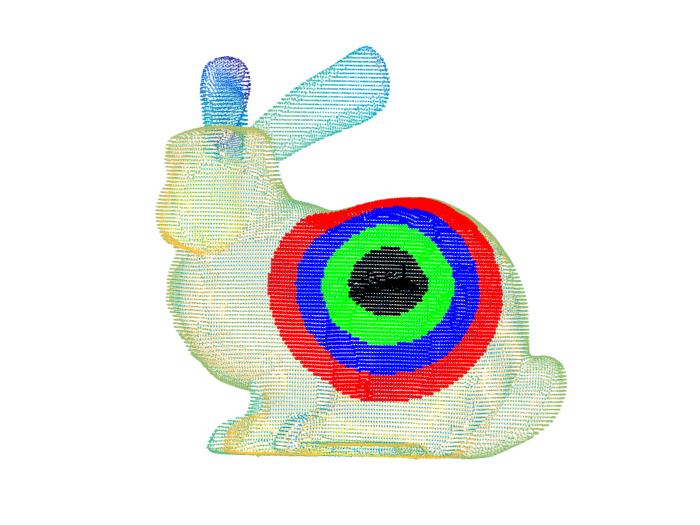}};
			\node (c1) at (-1,0){\includegraphics[scale=0.15]{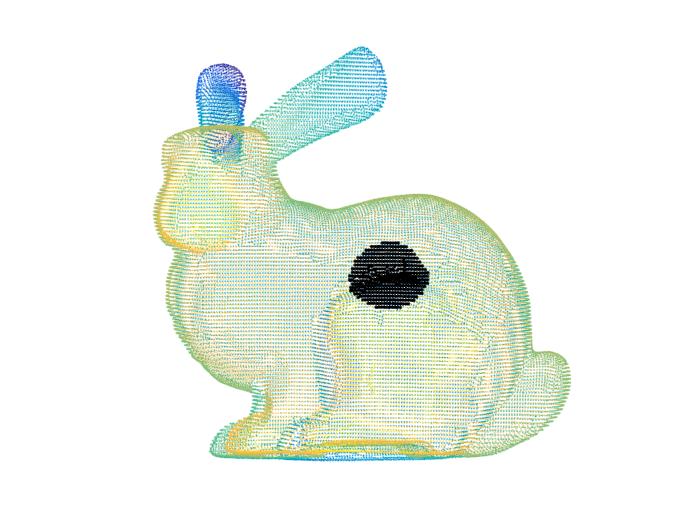}};
			\node (c2) at (3,0){\includegraphics[scale=0.15]{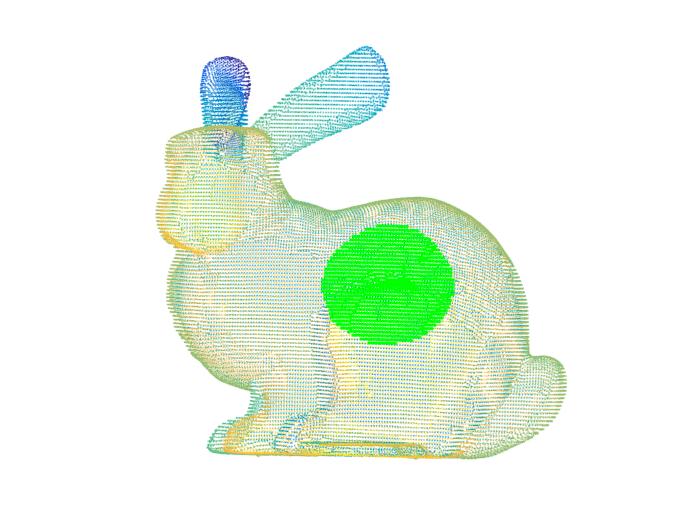}};
			\node (c3) at (7,0){\includegraphics[scale=0.15]{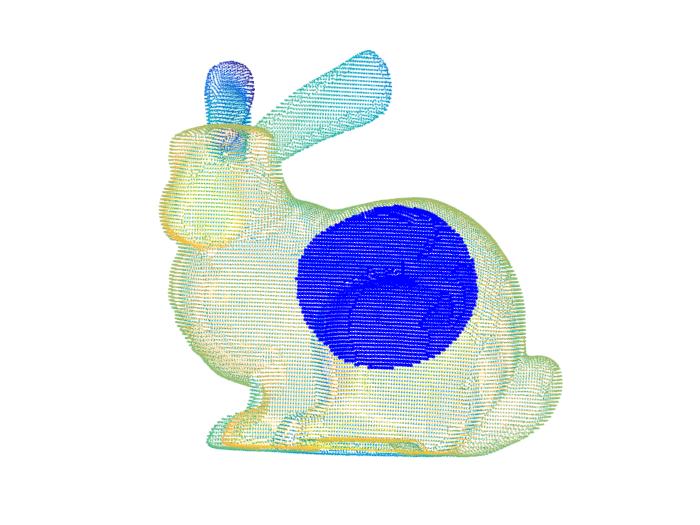}};
			\node (c4) at (11,0){\includegraphics[scale=0.15]{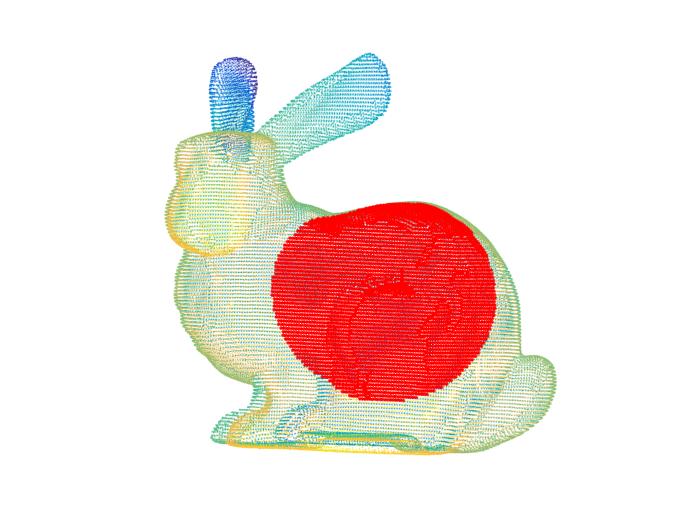}};
			\end{tikzpicture}
		\end{center}
		\caption{Multi-scale Receptive Fields}
	\end{figure}
	
	In order to obtain sufficient feature information and stabilize the network, the multi-scale receptive field strategy analogous to multi-heads mechanism is proposed. Unlike previous works , the sizes of receptive fields in our model are different for various branches.  Therefore, we concatenate $M$ independent SRFGAT module and generate a semantic feature with $M\times F'$ channels.
	\begin{equation}
	\tilde{x}_{i} =  \bigg{\|}_{m=1}^M\tilde{x}^{(m)}_i
	\end{equation}
	$\tilde{x}^{(m)}_i$is the receptive field feature of the $m$-th branch, $M$ is the total number of branches and $||$ is the concatenation operation over feature channels.
	
	\subsection{MRFGAT architecture}
	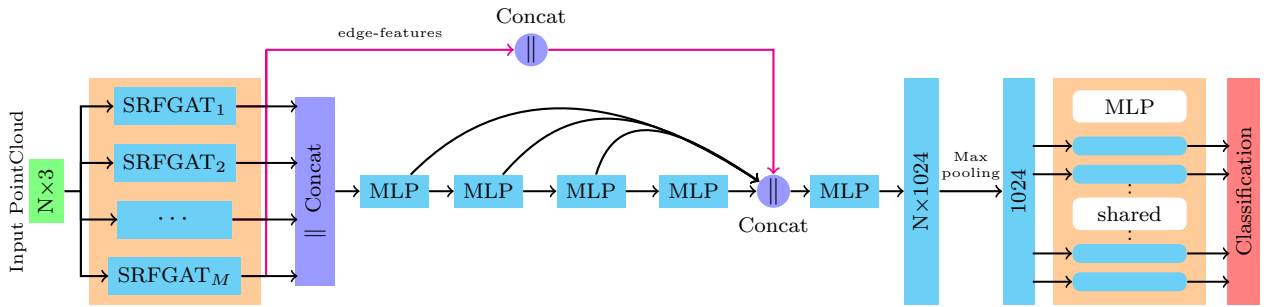
\begin{figure}[H]
		\begin{center}
			\begin{tikzpicture}[scale=0.75]
			\node [rectangle,fill=red!0,rotate=90] (in) at (0.0,3) {\scriptsize{Input PointCloud}};			
			\node [rectangle,fill=green!50,rotate=90] (input) at (0.5,3) {\footnotesize{N$\times$3}};
			\node [rectangle,fill=orange!40,minimum width=2.25cm, minimum height=3cm] (ATM) at (2.75,3) {};
			\coordinate (conect1) at (1.1,3);
			\node [rectangle,fill=cyan!50,minimum width=1.5cm, minimum height=0.5cm] (ATM1) at (2.75,4.5) {\footnotesize{SRFGAT$_1$}};
			\node [rectangle,fill=cyan!50,minimum width=1.5cm, minimum height=0.5cm] (ATM2) at (2.75,3.5) {\footnotesize{SRFGAT$_2$}};
			\node [rectangle,fill=cyan!50,minimum width=1.5cm, minimum height=0.5cm] (ATM3) at (2.75,2.5) {$\cdots$};
			\node [rectangle,fill=cyan!50,minimum width=1.5cm, minimum height=0.5cm] (ATM4) at (2.75,1.5) {\footnotesize{SRFGAT$_M$}};
			
			\draw [-,color=black,line width=0.8pt] (input)-- (conect1);
			\draw [->,color=black,line width=0.8pt] (conect1)|- (ATM1);
			\draw [->,color=black,line width=0.8pt] (conect1)|- (ATM2);
			\draw [->,color=black,line width=0.8pt] (conect1)|- (ATM3);
			\draw [->,color=black,line width=0.8pt] (conect1)|- (ATM4);
			
			\coordinate (connect2) at (4.35,5.5);	
			\draw [-,color=magenta,line width=0.8pt] (ATM1)-| (connect2);
			\draw [-,color=magenta,line width=0.8pt] (ATM2)-| (connect2);
			\draw [-,color=magenta,line width=0.8pt] (ATM3)-| (connect2);
			\draw [-,color=magenta,line width=0.8pt] (ATM4)-| (connect2);
			
			\node[circle, fill=blue!40, inner sep=0.5pt,label=above:\footnotesize{Concat}] (otimes1) at (9.0,5.5) {$\|$};
			\draw [->,color=magenta,line width=0.8pt] (connect2)--node [above, color=black] {\tiny{edge-features}} (otimes1);
			
			\node [rectangle,fill=blue!40,minimum width=2.5cm, minimum height=0.2cm,rotate=90] (concat) at (5.21,3) {\footnotesize{$\|$~~Concat}};
			\coordinate (pc1) at (4.9,4.5);
			\coordinate (pc2) at (4.9,3.5);
			\coordinate (pc3) at (4.9,2.5);
			\coordinate (pc4) at (4.9,1.5);
			
			\draw [->,color=black,line width=0.75pt] (ATM1)-- (pc1);
			\draw [->,color=black,line width=0.75pt] (ATM2)-- (pc2);
			\draw [->,color=black,line width=0.75pt] (ATM3)-- (pc3);
			\draw [->,color=black,line width=0.75pt] (ATM4)-- (pc4);
			
			\node [rectangle,fill=cyan!50,minimum width=0.75cm, minimum height=0.25cm] (MLP1) at (6.6,3) {\footnotesize{MLP}};	
			\node [rectangle,fill=cyan!50,minimum width=0.75cm, minimum height=0.25cm] (MLP2) at (8.25,3) {\footnotesize{MLP}};	
			\node [rectangle,fill=cyan!50,minimum width=0.75cm, minimum height=0.25cm] (MLP3) at (10.05,3) {\footnotesize{MLP}};
			\node [rectangle,fill=cyan!50,minimum width=0.75cm, minimum height=0.25cm] (MLP4) at (11.85,3) {\footnotesize{MLP}};	
			\draw [->,color=black,line width=0.75pt] (concat)-- (MLP1);	
			\draw [->,color=black,line width=0.75pt] (MLP1)-- (MLP2);
			\draw [->,color=black,line width=0.75pt] (MLP2)-- (MLP3);
			\draw [->,color=black,line width=0.75pt] (MLP3)-- (MLP4);
			
			\node[circle, fill=blue!40, inner sep=0.5pt,label=below:\footnotesize{Concat}] (otimes2) at (13.25,3) {$\|$};
			\draw [->,color=black,line width=0.75pt] (MLP4)-- (otimes2);
			\draw [->,color=magenta,line width=0.8pt] (otimes1)-| (otimes2);
			\draw[line width=0.8pt,,color=black, ->] (MLP1) ..controls(8.75,5.2) and (11.0,4.5).. (otimes2);
			\draw[line width=0.8pt,,color=black, ->] (MLP2) ..controls(9.75,4.75) and (11.0,4.5).. (otimes2);
			\draw[line width=0.8pt,,color=black, ->] (MLP3) ..controls(10.5,4.5) and (11.5,4.2).. (otimes2);
			
			\node [rectangle,fill=cyan!50,minimum width=0.75cm, minimum height=0.25cm] (MLP5) at (14.5,3) {\footnotesize {MLP}};
			
			\draw [->,color=black,line width=0.75pt] (otimes2)-- (MLP5);
			
			\node [rectangle,fill=cyan!50,minimum width=3cm, minimum height=0.2cm,rotate=90] (o1) at (15.85,3) {\footnotesize{N$\times$1024}};
			
			\draw [->,color=black,line width=0.75pt] (MLP5) --  (o1);	
			
			\node [rectangle,fill=cyan!50,minimum width=3cm, minimum height=0.2cm,rotate=90] (1024) at (17.55,3) {\footnotesize{1024}};
			\coordinate (po2-1) at (17.8,4.5);
			\coordinate (po2-2) at (17.8,3.8);
			\coordinate (po2-3) at (17.8,3.3);
			\coordinate (po2-4) at (17.8,2.6);
			\coordinate (po2-5) at (17.8,1.9);
			\coordinate (po2-6) at (17.8,1.4);
			
			\draw [->,color=black,line width=0.8pt] (o1)--node [above] {\tiny{$\begin{matrix} \text{Max}\\\text{pooling} \end{matrix}$}} (1024);

			\node [rectangle,fill=orange!40,minimum width=2.0cm, minimum height=3cm] (MLP-union) at (19.5,3) {};
			\node [rectangle,rounded corners=1mm,fill=cyan!0,minimum width=1.5cm, minimum height=0.25cm] (mlp1) at (19.5,4.5) {\footnotesize{MLP}};
			\node [rectangle,rounded corners=1mm, fill=cyan!50,minimum width=1.5cm, minimum height=0.25cm] (mlp2) at (19.5,3.8) {};
			\node [rectangle,rounded corners=1mm,fill=cyan!50,minimum width=1.5cm, minimum height=0.25cm] (mlp3) at (19.5,3.3) {};
			\node [rectangle,rounded corners=1mm,fill=cyan!0,minimum width=1.5cm, minimum height=0.25cm] (mlp4) at (19.5,2.6) {\footnotesize{shared}};
			\node [rectangle,rounded corners=1mm,fill=cyan!50,minimum width=1.5cm, minimum height=0.25cm] (mlp5) at (19.5,1.9) {};
			\node [rectangle,rounded corners=1mm,fill=cyan!50,minimum width=1.5cm, minimum height=0.25cm] (mlp6) at (19.5,1.4) {};
			
			\draw [->,color=black,line width=0.75pt] (po2-2)-- (mlp2);
			\draw [->,color=black,line width=0.75pt] (po2-3)-- (mlp3);
			\draw [->,color=black,line width=0.75pt] (po2-5)-- (mlp5);
			\draw [->,color=black,line width=0.75pt] (po2-6)-- (mlp6);
			\draw [dotted,color=black,line width=0.75pt] (mlp3)-- (mlp4);
			\draw [dotted,color=black,line width=0.75pt] (mlp4)-- (mlp5);
			
			\node [rectangle,fill=red!50,minimum width=3cm, minimum height=0.2cm,rotate=90] (o3) at (21.5,3) {\footnotesize{Classification}};
			\coordinate (po3-1) at (21.25,4.5);
			\coordinate (po3-2) at (21.25,3.8);
			\coordinate (po3-3) at (21.25,3.3);
			\coordinate (po3-4) at (21.25,2.6);
			\coordinate (po3-5) at (21.25,1.9);
			\coordinate (po3-6) at (21.25,1.4);
			
			\draw [<-,color=black,line width=0.75pt] (po3-2)-- (mlp2);
			\draw [<-,color=black,line width=0.75pt] (po3-3)-- (mlp3);
			\draw [<-,color=black,line width=0.75pt] (po3-5)-- (mlp5);
			\draw [<-,color=black,line width=0.75pt] (po3-6)-- (mlp6);		
			\end{tikzpicture}
		\end{center}
		\caption{The architecture of classification. In this framework, it takes $N$ points as input and applies $M$ individual SRFGAT-modules to obtain multi-attention features on  multi local graphs, then the output features are recast by means of five shared MLP layers and attention pooling layer, respectively. Finally, a shared full-connected layer is employed to form a global feature, then obtained classification scores for c categories.}
		\label{myNet}
	\end{figure}
	
	Our MRFGAT model shown in Figure \ref{myNet} considers shape classification task for point cloud. The architecture is similar to PointNet \cite{charles2017pointnet}. However, there are three main differences between the architectures. Firstly, according to the analyses of LinkDGCNN model, we remove the transformation network which is used in many architectures such as PointNet, DGCNN and GAPNet etc..  Secondly, instead of only processing individual points of point-cloud, we also exploit local features by a SRFGAP-Layer before the stacked MLP layers. Thirdly, an attention pooling layer is used to obtain local signature that is connected to the intermediate layer for capturing a global descriptor.	In addition, we aggregate individually the original edge-feature of every SRFGAP channel, then obtain a local features which can enhance the semantic feature of MRFGAT.

	\section{Experiments}\label{sec:04}
	In this section, we evaluate our MRFGAT model on 3D point cloud analysis for the classification tasks. To demonstrate effectiveness of our model,  we then compare the performance for our model to recent state-of-the-art methods and perform ablation study to investigate different design variations.
	
	\subsection{Classification}
	\textbf{Dataset}. We demonstrate the feasibility and effectiveness of our model on the ModelNet10 and ModelNet40 benchmarks \cite{wu20153d} for shape classification. The ModelNet40 dataset contains 12,311 meshed CAD models that are classified to 40 man-made categories. The ModelNet40 was separate 9843 models for training and 2468 models for testing. ModelNet10 contains 4,899 CAD models from 10 categories, it was split into 3991 training samples and 908 testing samples. Then we normalize the models in the unit sphere and uniformly sample 1,024 points over model surface. Besides, We further augment the training dataset by randomly rotating, scaling the point cloud and jittering the location of every point by means of Gaussian noise with zero mean and 0.01 standard deviation for all the models.
	
	\textbf{Implementation Details}. According to the analysis of Link-DGCNN model \cite{zhang2019linked}, we omit the spatial transformation network to align the point cloud to a canonical space. The network employs four SRFGAPLayer modules  with (8,16,16,24) channels to capture attention features, respectively. Then, four shared MLP layers with sizes (128, 64, 64, 64), respectively, followed it are used to aggregate the feature information. Next, the output features are  fed into a aggregation operation followed by MLP layer with 1024 neurons. In the end of network,  a max pooling operation and two full-connected layers (512, 256) are used to finally obtain the classification score. The training is done using Adam optimizer with mini-batch training (batch size of 16) and an initial learning rate of 0.001. The Relu activate function and Batch Normalization(BN) are also used in both SRFGAP module and MLP layer. At last, the network was implemented using TensorFlow and executed on server equipped with four NVIDIA GTX2080Ti.
	\begin{table}
		\centering
		\renewcommand\arraystretch{1.5}
		\begin{tabular}{|c|c|cc|cc|}
			\hline
			\multirow{2}*{Method} &\multirow{2}*{Points}&\multicolumn{2}{|c|}{ModelNet10}     & \multicolumn{2}{|c|}{ModelNet40}  \\ \cline{3-6}
			&                     &MA ($\%$)         &OA($\%$)          &MA($\%$)          &OA($\%$)       \\ \hline
			SO-Net \cite{li2018so-net}        &2048$\times$3        & 93.9             & 94.1             & 87.3             &90.9                 \\
			KD-Net\cite{klokov2017escape}     &1024$\times$3        & 93.5             & 94.0             & 88.5             &91.8                 \\
			PointNet\cite{charles2017pointnet}&1024$\times$3        & --               & --                 &86.0              & 89.2                 \\
			PointNet++\cite{qi2017pointnet}   &1024$\times$3        & --               & --                 & --               &90.7                 \\
			PointCNN\cite{pointCNN2018}       &1024$\times$3        & --               & --                 &88.1              &92.2                 \\
			DGCNN\cite{wang2019dynamic}       &1024$\times$3        &--                & --                 &90.2              &92.2                 \\
			PCNN\cite{atzmon2018point}        &1024$\times$3        & --               & --                 & --               &92.3                 \\
			GAPNet\cite{chen2019gapnet}       &1024$\times$3        & --               & --                 &89.7              &92.4                  \\ \hline
			Ours                              &1024$\times$3        &95.0              & 95.6               &90.1              &92.5                 \\
			\hline
		\end{tabular}
		\caption{Classification results on ModelNet10 and ModelNet40. MA represents mean per-class accuracy, and the per-class accuracy is the ratio of the number of correct classifications to that of the objects in a class. OA denotes the overall accuracy, which is the ratio of the number of overall correct classifications to that of overall objects. }
		\label{TT01}
	\end{table}
	
	\textbf{Results}. Table \ref{TT01} list the  results of our method and several recent state-of-the-art works. The methods listed in Table \ref{TT01}  have one thing in common. The input is only raw point cloud with 3D coordinates $(x_i, y_i, z_i)$. Based on this results, we can conclude that our model performs better than other methods and obtains wonderful performance on both the ModelNet10 and ModelNet40 benchmark. For ModelNet 10, our model is superior to that of SO-Net and KD-Net. Comparing to other point-based methods, the performance for our model is only a little weaker than that of DGCNN in terms of MA on ModelNet 40. But it outperforms the previous state-of-the-art model GAPNet by 0.1 $\%$ accuracy in terms of OA.  These phenomena show that the strategy employing local and global features in different receptive fields is efficient, it will help us to capture the prominent semantic feature for point cloud. And in our model, since we introduce the structure of the data by providing the local interconnection between points and explore graph features from different scale field levels by the localized graph convolutional layers, it guarantees the exploration of more distinctive latent representations for each object class.

	\section{Conclusion}\label{sec:05}
	Current advances in graph convolutional networks have led to better performances in varied 3D computer vision tasks. This has motivated us to leverage GCNs for the task of 3D vehicle detection, and demonstrate their effectiveness for vehicle detection. We introduced an novel MRFGAT-based modules for point feature and context aggregation both within and between proposals. Making use of different receptive feild and attention strategy, the pipeline MRFGAT can capture more fine feaures of point clouds for classification task and other vision tasks. We showed comparable results with recent works, and show it achieves state-of-the-art performance on the dataset ModelNet. Based on the state-of-the-art Graph Convolution Networks(GCN) for semantic segmentation in point cloud, it would be interesting to introduced an efficient GCN-like operation for our model to address unstructured data in the future.
	
	
%

\begin{thebibliography}{50}
\expandafter\ifx\csname natexlab\endcsname\relax\def\natexlab#1{#1}\fi
\providecommand{\url}[1]{\texttt{#1}}
\providecommand{\href}[2]{#2}
\providecommand{\path}[1]{#1}
\providecommand{\DOIprefix}{doi:}
\providecommand{\ArXivprefix}{arXiv:}
\providecommand{\URLprefix}{URL: }
\providecommand{\Pubmedprefix}{pmid:}
\providecommand{\doi}[1]{\href{http://dx.doi.org/#1}{\path{#1}}}
\providecommand{\Pubmed}[1]{\href{pmid:#1}{\path{#1}}}
\providecommand{\bibinfo}[2]{#2}
\ifx\xfnm\relax \def\xfnm[#1]{\unskip,\space#1}\fi
\bibitem[{{Zhou} and {Tuzel}(2018)}]{zhou2018voxelnet}
\bibinfo{author}{Y.~{Zhou}}, \bibinfo{author}{O.~{Tuzel}},
\newblock \bibinfo{title}{{VoxelNet: End-to-End Learning for Point Cloud Based
  3D Object Detection}},
\newblock in: \bibinfo{booktitle}{2018 IEEE/CVF Conference on Computer Vision
  and Pattern Recognition}, \bibinfo{year}{2018}, pp.
  \bibinfo{pages}{4490--4499}.
\bibitem[{{Qi} et~al.(2018){Qi}, {Liu}, {Wu}, {Su}, and
  {Guibas}}]{qi2018frustum}
\bibinfo{author}{C.~R. {Qi}}, \bibinfo{author}{W.~{Liu}},
  \bibinfo{author}{C.~{Wu}}, \bibinfo{author}{H.~{Su}}, \bibinfo{author}{L.~J.
  {Guibas}},
\newblock \bibinfo{title}{{Frustum PointNets for 3D Object Detection from RGB-D
  Data}},
\newblock in: \bibinfo{booktitle}{2018 IEEE/CVF Conference on Computer Vision
  and Pattern Recognition}, \bibinfo{year}{2018}, pp.
  \bibinfo{pages}{918--927}.
\bibitem[{{Ku} et~al.(2017){Ku}, {Mozifian}, {Lee}, {Harakeh}, and
  {Waslander}}]{ku2017joint}
\bibinfo{author}{J.~{Ku}}, \bibinfo{author}{M.~{Mozifian}},
  \bibinfo{author}{J.~{Lee}}, \bibinfo{author}{A.~{Harakeh}},
  \bibinfo{author}{S.~{Waslander}},
\newblock \bibinfo{title}{{Joint 3D Proposal Generation and Object Detection
  from View Aggregation}},
\newblock \bibinfo{journal}{arXiv preprint arXiv:1712.02294}
  (\bibinfo{year}{2017}).
\bibitem[{{Wang} et~al.(2018){Wang}, {Suo}, {Ma}, {Pokrovsky}, and
  {Urtasun}}]{wang2018deep}
\bibinfo{author}{S.~{Wang}}, \bibinfo{author}{S.~{Suo}}, \bibinfo{author}{W.-C.
  {Ma}}, \bibinfo{author}{A.~{Pokrovsky}}, \bibinfo{author}{R.~{Urtasun}},
\newblock \bibinfo{title}{{Deep Parametric Continuous Convolutional Neural
  Networks}},
\newblock in: \bibinfo{booktitle}{2018 IEEE/CVF Conference on Computer Vision
  and Pattern Recognition}, \bibinfo{year}{2018}, pp.
  \bibinfo{pages}{2589--2597}.
\bibitem[{{Liang} et~al.(2018){Liang}, {Yang}, {Wang}, and
  {Urtasun}}]{liang2018deep}
\bibinfo{author}{M.~{Liang}}, \bibinfo{author}{B.~{Yang}},
  \bibinfo{author}{S.~{Wang}}, \bibinfo{author}{R.~{Urtasun}},
\newblock \bibinfo{title}{{Deep Continuous Fusion for Multi-Sensor 3D Object
  Detection}},
\newblock in: \bibinfo{booktitle}{Proceedings of the European Conference on
  Computer Vision (ECCV)}, \bibinfo{year}{2018}, pp. \bibinfo{pages}{663--678}.
\bibitem[{{Biswas} and {Veloso}(2012)}]{biswas2012depth}
\bibinfo{author}{J.~{Biswas}}, \bibinfo{author}{M.~{Veloso}},
\newblock \bibinfo{title}{{Depth camera based indoor mobile robot localization
  and navigation}},
\newblock in: \bibinfo{booktitle}{2012 IEEE International Conference on
  Robotics and Automation}, volume \bibinfo{volume}{2012},
  \bibinfo{year}{2012}, pp. \bibinfo{pages}{1697--1702}.
\bibitem[{{Zhu} et~al.(2017){Zhu}, {Mottaghi}, {Kolve}, {Lim}, {Gupta},
  {Fei-Fei}, and {Farhadi}}]{zhu2017target}
\bibinfo{author}{Y.~{Zhu}}, \bibinfo{author}{R.~{Mottaghi}},
  \bibinfo{author}{E.~{Kolve}}, \bibinfo{author}{J.~J. {Lim}},
  \bibinfo{author}{A.~{Gupta}}, \bibinfo{author}{L.~{Fei-Fei}},
  \bibinfo{author}{A.~{Farhadi}},
\newblock \bibinfo{title}{{Target-driven visual navigation in indoor scenes
  using deep reinforcement learning}},
\newblock in: \bibinfo{booktitle}{2017 IEEE International Conference on
  Robotics and Automation (ICRA)}, \bibinfo{year}{2017}, pp.
  \bibinfo{pages}{3357--3364}.
\bibitem[{Golovinskiy et~al.(2009)Golovinskiy, Kim, and
  Funkhouser}]{Golovinskiy2009Shape}
\bibinfo{author}{A.~Golovinskiy}, \bibinfo{author}{V.~G. Kim},
  \bibinfo{author}{T.~Funkhouser},
\newblock \bibinfo{title}{{Shape-based recognition of 3D point clouds in urban
  environments}},
\newblock in: \bibinfo{booktitle}{Computer Vision, 2009 IEEE 12th International
  Conference on}, \bibinfo{year}{2009}.
\bibitem[{{Mitra} et~al.(2004){Mitra}, {Gelfand}, {Pottmann}, and
  {Guibas}}]{mitra2004registration}
\bibinfo{author}{N.~J. {Mitra}}, \bibinfo{author}{N.~{Gelfand}},
  \bibinfo{author}{H.~{Pottmann}}, \bibinfo{author}{L.~{Guibas}},
\newblock \bibinfo{title}{{Registration of point cloud data from a geometric
  optimization perspective}},
\newblock in: \bibinfo{booktitle}{Proceedings of the 2004 Eurographics/ACM
  SIGGRAPH symposium on Geometry processing}, \bibinfo{year}{2004}, pp.
  \bibinfo{pages}{22--31}.
\bibitem[{{Vosselman} and {Dijkman}(2001)}]{vosselman20013d}
\bibinfo{author}{G.~{Vosselman}}, \bibinfo{author}{S.~{Dijkman}},
\newblock \bibinfo{title}{{3D building model reconstruction from point clouds
  and ground plans}},
\newblock \bibinfo{journal}{ISPRS Workshop: land surface mapping and
  characterization using laser altimetry}  (\bibinfo{year}{2001})
  \bibinfo{pages}{37--43}.
\bibitem[{Rusu et~al.(2009)Rusu, Blodow, and Beetz}]{Rusu2009Fast}
\bibinfo{author}{R.~B. Rusu}, \bibinfo{author}{N.~Blodow},
  \bibinfo{author}{M.~Beetz},
\newblock \bibinfo{title}{{Fast Point Feature Histograms (FPFH) for 3D
  registration}},
\newblock in: \bibinfo{booktitle}{Robotics and Automation, 2009. ICRA '09. IEEE
  International Conference on}, \bibinfo{year}{2009}.
\bibitem[{Tombari et~al.(2010)Tombari, Salti, and Stefano}]{Tombari2010Unique}
\bibinfo{author}{F.~Tombari}, \bibinfo{author}{S.~Salti},
  \bibinfo{author}{L.~D. Stefano},
\newblock \bibinfo{title}{{Unique Signatures of Histograms for Local Surface
  Description}},
\newblock in: \bibinfo{booktitle}{Computer Vision - ECCV 2010, 11th European
  Conference on Computer Vision, Heraklion, Crete, Greece, September 5-11,
  2010, Proceedings, Part III}, \bibinfo{year}{2010}.
\bibitem[{{Charles} et~al.(2017){Charles}, {Su}, {Kaichun}, and
  {Guibas}}]{charles2017pointnet}
\bibinfo{author}{R.~Q. {Charles}}, \bibinfo{author}{H.~{Su}},
  \bibinfo{author}{M.~{Kaichun}}, \bibinfo{author}{L.~J. {Guibas}},
\newblock \bibinfo{title}{{PointNet: Deep Learning on Point Sets for 3D
  Classification and Segmentation}},
\newblock in: \bibinfo{booktitle}{2017 IEEE Conference on Computer Vision and
  Pattern Recognition (CVPR)}, \bibinfo{year}{2017}, pp.
  \bibinfo{pages}{77--85}.
\bibitem[{Yangyan~{Li} and {Chen}(2018)}]{pointCNN2018}
\bibinfo{author}{M.~S. W. W. X.~D. Yangyan~{Li}, Rui~{Bu}},
  \bibinfo{author}{B.~{Chen}},
\newblock \bibinfo{title}{{PointCNN: convolution on $\mathcal{X}$-transformed
  points}},
\newblock in: \bibinfo{booktitle}{Advances in Neural Information Processing
  Systems(NIPS)}, \bibinfo{year}{2018}, pp. \bibinfo{pages}{828--838}.
\bibitem[{{Wang} et~al.(2019){Wang}, {Sun}, {Liu}, {Sarma}, {Bronstein}, and
  {Solomon}}]{wang2019dynamic}
\bibinfo{author}{Y.~{Wang}}, \bibinfo{author}{Y.~{Sun}},
  \bibinfo{author}{Z.~{Liu}}, \bibinfo{author}{S.~E. {Sarma}},
  \bibinfo{author}{M.~M. {Bronstein}}, \bibinfo{author}{J.~M. {Solomon}},
\newblock \bibinfo{title}{{Dynamic Graph CNN for Learning on Point Clouds}},
\newblock \bibinfo{journal}{ACM Transactions on Graphics} \bibinfo{volume}{38}
  (\bibinfo{year}{2019}) \bibinfo{pages}{146}.
\bibitem[{{Zhao} et~al.(2019){Zhao}, {Jiang}, {Fu}, and
  {Jia}}]{zhao2019pointweb}
\bibinfo{author}{H.~{Zhao}}, \bibinfo{author}{L.~{Jiang}},
  \bibinfo{author}{C.-W. {Fu}}, \bibinfo{author}{J.~{Jia}},
\newblock \bibinfo{title}{{PointWeb: Enhancing Local Neighborhood Features for
  Point Cloud Processing}},
\newblock in: \bibinfo{booktitle}{2019 IEEE/CVF Conference on Computer Vision
  and Pattern Recognition (CVPR)}, \bibinfo{year}{2019}, pp.
  \bibinfo{pages}{5565--5573}.
\bibitem[{{Thomas} et~al.(2019){Thomas}, {Qi}, {Deschaud}, {Marcotegui},
  {Goulette}, and {Guibas}}]{thomas2019kpconv}
\bibinfo{author}{H.~{Thomas}}, \bibinfo{author}{C.~R. {Qi}},
  \bibinfo{author}{J.-E. {Deschaud}}, \bibinfo{author}{B.~{Marcotegui}},
  \bibinfo{author}{F.~{Goulette}}, \bibinfo{author}{L.~J. {Guibas}},
\newblock \bibinfo{title}{{KPConv: Flexible and Deformable Convolution for
  Point Clouds}},
\newblock in: \bibinfo{booktitle}{Proceedings of the IEEE International
  Conference on Computer Vision}, \bibinfo{year}{2019}, pp.
  \bibinfo{pages}{6411--6420}.
\bibitem[{{Qi} et~al.(2017){Qi}, {Yi}, {Su}, and {Guibas}}]{qi2017pointnet}
\bibinfo{author}{C.~R. {Qi}}, \bibinfo{author}{L.~{Yi}},
  \bibinfo{author}{H.~{Su}}, \bibinfo{author}{L.~J. {Guibas}},
\newblock \bibinfo{title}{{PointNet++: Deep Hierarchical Feature Learning on
  Point Sets in a Metric Space}},
\newblock in: \bibinfo{booktitle}{Advances in Neural Information Processing
  Systems}, \bibinfo{year}{2017}, pp. \bibinfo{pages}{5099--5108}.
\bibitem[{{Atzmon} et~al.(2018){Atzmon}, {Maron}, and
  {Lipman}}]{atzmon2018point}
\bibinfo{author}{M.~{Atzmon}}, \bibinfo{author}{H.~{Maron}},
  \bibinfo{author}{Y.~{Lipman}},
\newblock \bibinfo{title}{{Point Convolutional Neural Networks by Extension
  Operators}},
\newblock \bibinfo{journal}{international conference on computer graphics and
  interactive techniques} \bibinfo{volume}{37} (\bibinfo{year}{2018})
  \bibinfo{pages}{71}.
\bibitem[{Jiang et~al.(2018)Jiang, Wu, Zhao, Zhao, and Lu}]{jiang2018pointsift}
\bibinfo{author}{M.~Jiang}, \bibinfo{author}{Y.~Wu}, \bibinfo{author}{T.~Zhao},
  \bibinfo{author}{Z.~Zhao}, \bibinfo{author}{C.~Lu},
\newblock \bibinfo{title}{{PointSIFT: A SIFT-like Network Module for 3D Point
  Cloud Semantic Segmentation.}},
\newblock \bibinfo{journal}{Computer Vision and Pattern Recognition}
  (\bibinfo{year}{2018}).
\bibitem[{{Gori} et~al.(2005){Gori}, {Monfardini}, and {Scarselli}}]{2005anew}
\bibinfo{author}{M.~{Gori}}, \bibinfo{author}{G.~{Monfardini}},
  \bibinfo{author}{F.~{Scarselli}},
\newblock \bibinfo{title}{{A new model for learning in graph domains}},
\newblock in: \bibinfo{booktitle}{Proceedings. 2005 IEEE International Joint
  Conference on Neural Networks, 2005.}, volume~\bibinfo{volume}{2},
  \bibinfo{year}{2005}, pp. \bibinfo{pages}{729--734}.
\bibitem[{{Scarselli} et~al.(2009){Scarselli}, {Gori}, {Tsoi}, {Hagenbuchner},
  and {Monfardini}}]{2009graph}
\bibinfo{author}{F.~{Scarselli}}, \bibinfo{author}{M.~{Gori}},
  \bibinfo{author}{A.~C. {Tsoi}}, \bibinfo{author}{M.~{Hagenbuchner}},
  \bibinfo{author}{G.~{Monfardini}},
\newblock \bibinfo{title}{{The Graph Neural Network Model}},
\newblock \bibinfo{journal}{IEEE Transactions on Neural Networks}
  \bibinfo{volume}{20} (\bibinfo{year}{2009}) \bibinfo{pages}{61--80}.
\bibitem[{{Zhang} et~al.(2019){Zhang}, {Hao}, {Wang}, de~{Silva}, and
  {Fu}}]{zhang2019linked}
\bibinfo{author}{K.~{Zhang}}, \bibinfo{author}{M.~{Hao}},
  \bibinfo{author}{J.~{Wang}}, \bibinfo{author}{C.~W. de~{Silva}},
  \bibinfo{author}{C.~{Fu}},
\newblock \bibinfo{title}{{Linked Dynamic Graph CNN:Learning on Point Cloud via
  Linking Hierarchical Features}},
\newblock \bibinfo{journal}{arXiv preprint arXiv:1904.10014}
  (\bibinfo{year}{2019}).
\bibitem[{Te et~al.(2018)Te, Hu, Guo, and Zheng}]{te2018rgcnn}
\bibinfo{author}{G.~Te}, \bibinfo{author}{W.~Hu}, \bibinfo{author}{Z.~Guo},
  \bibinfo{author}{A.~Zheng},
\newblock \bibinfo{title}{{RGCNN: Regularized Graph CNN for Point Cloud
  Segmentation}},
\newblock \bibinfo{journal}{Computer Vision and Pattern Recognition}
  (\bibinfo{year}{2018}).
\bibitem[{Gao et~al.(2019)Gao, Hu, and Guo}]{gao2019exploring}
\bibinfo{author}{X.~Gao}, \bibinfo{author}{W.~Hu}, \bibinfo{author}{Z.~Guo},
\newblock \bibinfo{title}{{Exploring Structure-Adaptive Graph Learning for
  Robust Semi-Supervised Classification.}},
\newblock \bibinfo{journal}{arXiv: Learning}  (\bibinfo{year}{2019}).
\bibitem[{Lu et~al.(2020)Lu, Chen, Xie, and Luo}]{lu2020pointngcnn}
\bibinfo{author}{Q.~Lu}, \bibinfo{author}{C.~Chen}, \bibinfo{author}{W.~Xie},
  \bibinfo{author}{Y.~Luo},
\newblock \bibinfo{title}{{PointNGCNN: Deep convolutional networks on 3D point
  clouds with neighborhood graph filters}},
\newblock \bibinfo{journal}{Computers and Graphics} \bibinfo{volume}{86}
  (\bibinfo{year}{2020}) \bibinfo{pages}{42--51}.
\bibitem[{{Vaswani} et~al.(2017){Vaswani}, {Shazeer}, {Parmar}, {Uszkoreit},
  {Jones}, {Gomez}, {Kaiser}, and {Polosukhin}}]{vaswani2017attention}
\bibinfo{author}{A.~{Vaswani}}, \bibinfo{author}{N.~{Shazeer}},
  \bibinfo{author}{N.~{Parmar}}, \bibinfo{author}{J.~{Uszkoreit}},
  \bibinfo{author}{L.~{Jones}}, \bibinfo{author}{A.~N. {Gomez}},
  \bibinfo{author}{L.~{Kaiser}}, \bibinfo{author}{I.~{Polosukhin}},
\newblock \bibinfo{title}{Attention is all you need},
\newblock \bibinfo{journal}{Neural Information Processing Systems}
  (\bibinfo{year}{2017}) \bibinfo{pages}{6000--6010}.
\bibitem[{Mnih et~al.(2014)Mnih, Heess, Graves, and
  Kavukcuoglu}]{mnih2014recurrent}
\bibinfo{author}{V.~Mnih}, \bibinfo{author}{N.~Heess},
  \bibinfo{author}{A.~Graves}, \bibinfo{author}{K.~Kavukcuoglu},
\newblock \bibinfo{title}{{Recurrent Models of Visual Attention}},
\newblock \bibinfo{journal}{Neural Information Processing Systems}
  (\bibinfo{year}{2014}) \bibinfo{pages}{2204--2212}.
\bibitem[{{Velikovi} et~al.(2017){Veliovi}, {Cucurull}, {Casanova},
  {Romero}, {Li}, and {Bengio}}]{GAT2017graph}
\bibinfo{author}{P.~{Velikovi}}, \bibinfo{author}{G.~{Cucurull}},
  \bibinfo{author}{A.~{Casanova}}, \bibinfo{author}{A.~{Romero}},
  \bibinfo{author}{P.~{Li}}, \bibinfo{author}{Y.~{Bengio}},
\newblock \bibinfo{title}{{Graph Attention Networks}},
\newblock \bibinfo{journal}{arXiv preprint arXiv:1710.10903}
  (\bibinfo{year}{2017}).
\bibitem[{{Chen} et~al.(2019){Chen}, {Fragonara}, and
  {Tsourdos}}]{chen2019gapnet}
\bibinfo{author}{C.~{Chen}}, \bibinfo{author}{L.~Z. {Fragonara}},
  \bibinfo{author}{A.~{Tsourdos}},
\newblock \bibinfo{title}{{GAPNet: Graph Attention based Point Neural Network
  for Exploiting Local Feature of Point Cloud.}},
\newblock \bibinfo{journal}{arXiv preprint arXiv:1905.08705}
  (\bibinfo{year}{2019}).
\bibitem[{Wang et~al.(2019)Wang, Huang, Hou, Zhang, and Shan}]{wang2019graph}
\bibinfo{author}{L.~Wang}, \bibinfo{author}{Y.~Huang},
  \bibinfo{author}{Y.~Hou}, \bibinfo{author}{S.~Zhang},
  \bibinfo{author}{J.~Shan},
\newblock \bibinfo{title}{{Graph Attention Convolution for Point Cloud Semantic
  Segmentation}},
\newblock \bibinfo{journal}{Computer Vision and Pattern Recognition}
  (\bibinfo{year}{2019}) \bibinfo{pages}{10296--10305}.
\bibitem[{Feng et~al.(2019)Feng, Zhang, Lin, Gilani, and Mian}]{feng2019point}
\bibinfo{author}{M.~Feng}, \bibinfo{author}{L.~Zhang},
  \bibinfo{author}{X.~Lin}, \bibinfo{author}{S.~Z. Gilani},
  \bibinfo{author}{A.~Mian},
\newblock \bibinfo{title}{{Point Attention Network for Semantic Segmentation of
  3D Point Clouds.}},
\newblock \bibinfo{journal}{Computer Vision and Pattern Recognition}
  (\bibinfo{year}{2019}).
\bibitem[{Kang and Ning(2019)}]{kang2019pyramnet}
\bibinfo{author}{Z.~Kang}, \bibinfo{author}{L.~Ning},
\newblock \bibinfo{title}{{PyramNet: Point Cloud Pyramid Attention Network and
  Graph Embedding Module for Classification and Segmentation.}},
\newblock \bibinfo{journal}{Computer Vision and Pattern Recognition}
  (\bibinfo{year}{2019}).
\bibitem[{{Niepert} et~al.(2016){Niepert}, {Ahmed}, and
  {Kutzkov}}]{niepert2016learning}
\bibinfo{author}{M.~{Niepert}}, \bibinfo{author}{M.~{Ahmed}},
  \bibinfo{author}{K.~{Kutzkov}},
\newblock \bibinfo{title}{{Learning convolutional neural networks for graphs}},
\newblock in: \bibinfo{booktitle}{ICML'16 Proceedings of the 33rd International
  Conference on International Conference on Machine Learning - Volume 48},
  \bibinfo{year}{2016}, pp. \bibinfo{pages}{2014--2023}.
\bibitem[{Li et~al.(2018)Li, Chen, and Lee}]{li2018so-net}
\bibinfo{author}{J.~Li}, \bibinfo{author}{B.~M. Chen}, \bibinfo{author}{G.~H.
  Lee},
\newblock \bibinfo{title}{{SO-Net: Self-Organizing Network for Point Cloud
  Analysis}},
\newblock \bibinfo{journal}{Computer Vision and Pattern Recognition}
  (\bibinfo{year}{2018}) \bibinfo{pages}{9397--9406}.
\bibitem[{{Yu} et~al.(2018){Yu}, {Li}, {Fu}, {Cohen-Or}, and {Heng}}]{yu2018pu}
\bibinfo{author}{L.~{Yu}}, \bibinfo{author}{X.~{Li}}, \bibinfo{author}{C.-W.
  {Fu}}, \bibinfo{author}{D.~{Cohen-Or}}, \bibinfo{author}{P.-A. {Heng}},
\newblock \bibinfo{title}{{PU-Net: Point Cloud Upsampling Network}},
\newblock in: \bibinfo{booktitle}{2018 IEEE/CVF Conference on Computer Vision
  and Pattern Recognition}, \bibinfo{year}{2018}, pp.
  \bibinfo{pages}{2790--2799}.
\bibitem[{{Klokov} and {Lempitsky}(2017)}]{klokov2017escape}
\bibinfo{author}{R.~{Klokov}}, \bibinfo{author}{V.~{Lempitsky}},
\newblock \bibinfo{title}{Escape from cells: Deep kd-networks for the
  recognition of 3d point cloud models},
\newblock in: \bibinfo{booktitle}{2017 IEEE International Conference on
  Computer Vision (ICCV)}, \bibinfo{year}{2017}, pp. \bibinfo{pages}{863--872}.
\bibitem[{{You} et~al.(2018){You}, {Lou}, {Liu}, {Tai}, {Ma}, {Lu}, and
  {Wang}}]{you2018pointwise}
\bibinfo{author}{Y.~{You}}, \bibinfo{author}{Y.~{Lou}},
  \bibinfo{author}{Q.~{Liu}}, \bibinfo{author}{Y.-W. {Tai}},
  \bibinfo{author}{L.~{Ma}}, \bibinfo{author}{C.~{Lu}},
  \bibinfo{author}{W.~{Wang}},
\newblock \bibinfo{title}{{Pointwise Rotation-Invariant Network with Adaptive
  Sampling and 3D Spherical Voxel Convolution.}},
\newblock \bibinfo{journal}{arXiv preprint arXiv:1811.09361}
  (\bibinfo{year}{2018}).
\bibitem[{{Wu} et~al.(2019){Wu}, {Qi}, and {Fuxin}}]{wu2019pointconv}
\bibinfo{author}{W.~{Wu}}, \bibinfo{author}{Z.~{Qi}},
  \bibinfo{author}{L.~{Fuxin}},
\newblock \bibinfo{title}{{PointConv: Deep Convolutional Networks on 3D Point
  Clouds}},
\newblock in: \bibinfo{booktitle}{2019 IEEE/CVF Conference on Computer Vision
  and Pattern Recognition (CVPR)}, \bibinfo{year}{2019}, pp.
  \bibinfo{pages}{9621--9630}.
\bibitem[{Zhao et~al.(2019)Zhao, Birdal, Deng, and Tombari}]{zhao20193d}
\bibinfo{author}{Y.~Zhao}, \bibinfo{author}{T.~Birdal},
  \bibinfo{author}{H.~Deng}, \bibinfo{author}{F.~Tombari},
\newblock \bibinfo{title}{{3D Point Capsule Networks}},
\newblock \bibinfo{journal}{arXiv: Computer Vision and Pattern Recognition}
  (\bibinfo{year}{2019}) \bibinfo{pages}{1009--1018}.
\bibitem[{Srivastava et~al.(2019)Srivastava, Goh, and
  Salakhutdinov}]{srivastava2019geometric}
\bibinfo{author}{N.~Srivastava}, \bibinfo{author}{H.~Goh},
  \bibinfo{author}{R.~Salakhutdinov},
\newblock \bibinfo{title}{{Geometric Capsule Autoencoders for 3D Point
  Clouds.}},
\newblock \bibinfo{journal}{arXiv: Learning}  (\bibinfo{year}{2019}).
\bibitem[{Cheraghian and Petersson(????)}]{cheraghian20193dcapsule}
\bibinfo{author}{A.~Cheraghian}, \bibinfo{author}{L.~Petersson},
\newblock \bibinfo{title}{3dcapsule: Extending the capsule architecture to
  classify 3d point clouds},
\newblock \bibinfo{journal}{workshop on applications of computer vision}
  (????) \bibinfo{pages}{1194--1202}.
\bibitem[{{Xu} et~al.(2018){Xu}, {Fan}, {Xu}, {Zeng}, and
  {Qiao}}]{xu2018spidercnn}
\bibinfo{author}{Y.~{Xu}}, \bibinfo{author}{T.~{Fan}},
  \bibinfo{author}{M.~{Xu}}, \bibinfo{author}{L.~{Zeng}},
  \bibinfo{author}{Y.~{Qiao}},
\newblock \bibinfo{title}{{SpiderCNN: Deep Learning on Point Sets with
  Parameterized Convolutional Filters}},
\newblock in: \bibinfo{booktitle}{Proceedings of the European Conference on
  Computer Vision (ECCV)}, \bibinfo{year}{2018}, pp. \bibinfo{pages}{87--102}.
\bibitem[{{Boscaini} et~al.(2015){Boscaini}, {Masci}, {Melzi}, {Bronstein},
  {Castellani}, and {Vandergheynst}}]{boscaini2015learning}
\bibinfo{author}{D.~{Boscaini}}, \bibinfo{author}{J.~{Masci}},
  \bibinfo{author}{S.~{Melzi}}, \bibinfo{author}{M.~M. {Bronstein}},
  \bibinfo{author}{U.~{Castellani}}, \bibinfo{author}{P.~{Vandergheynst}},
\newblock \bibinfo{title}{{Learning class-specific descriptors for deformable
  shapes using localized spectral convolutional networks}},
\newblock in: \bibinfo{booktitle}{Proceedings of the Eurographics Symposium on
  Geometry Processing}, volume~\bibinfo{volume}{34}, \bibinfo{year}{2015}, pp.
  \bibinfo{pages}{13--23}.
\bibitem[{{Yi} et~al.(2017){Yi}, {Su}, {Guo}, and {Guibas}}]{yi2017syncspeccnn}
\bibinfo{author}{L.~{Yi}}, \bibinfo{author}{H.~{Su}},
  \bibinfo{author}{X.~{Guo}}, \bibinfo{author}{L.~{Guibas}},
\newblock \bibinfo{title}{{SyncSpecCNN: Synchronized Spectral CNN for 3D Shape
  Segmentation}},
\newblock in: \bibinfo{booktitle}{2017 IEEE Conference on Computer Vision and
  Pattern Recognition (CVPR)}, \bibinfo{year}{2017}, pp.
  \bibinfo{pages}{6584--6592}.
\bibitem[{{Defferrard} et~al.(2016){Defferrard}, {Bresson}, and
  {Vandergheynst}}]{defferrard2016convolutional}
\bibinfo{author}{M.~{Defferrard}}, \bibinfo{author}{X.~{Bresson}},
  \bibinfo{author}{P.~{Vandergheynst}},
\newblock \bibinfo{title}{{Convolutional neural networks on graphs with fast
  localized spectral filtering}},
\newblock in: \bibinfo{booktitle}{NIPS'16 Proceedings of the 30th International
  Conference on Neural Information Processing Systems}, \bibinfo{year}{2016},
  pp. \bibinfo{pages}{3844--3852}.
\bibitem[{{Xiu} et~al.(2019){Xiu}, {Shinohara}, and
  {Matsuoka}}]{dynamicScale2019}
\bibinfo{author}{H.~{Xiu}}, \bibinfo{author}{T.~{Shinohara}},
  \bibinfo{author}{M.~{Matsuoka}},
\newblock \bibinfo{title}{Dynamic-scale graph convolutional network for
  semantic segmentation of 3d point cloud},
\newblock in: \bibinfo{booktitle}{2019 IEEE International Symposium on
  Multimedia (ISM)}, \bibinfo{year}{2019}, pp. \bibinfo{pages}{271--2717}.
\bibitem[{Verma et~al.(????)Verma, Boyer, and Verbeek}]{verma2018feastnet}
\bibinfo{author}{N.~Verma}, \bibinfo{author}{E.~Boyer},
  \bibinfo{author}{J.~Verbeek},
\newblock \bibinfo{title}{Feastnet: Feature-steered graph convolutions for 3d
  shape analysis},
\newblock \bibinfo{journal}{Computer Vision and Pattern Recognition}  (????)
  \bibinfo{pages}{2598--2606}.
\bibitem[{{Lee} et~al.(2018){Lee}, {Rossi}, {Kim}, {Ahmed}, and
  {Koh}}]{lee2018attention}
\bibinfo{author}{J.~B. {Lee}}, \bibinfo{author}{R.~A. {Rossi}},
  \bibinfo{author}{S.~{Kim}}, \bibinfo{author}{N.~K. {Ahmed}},
  \bibinfo{author}{E.~{Koh}},
\newblock \bibinfo{title}{Attention models in graphs: A survey},
\newblock \bibinfo{journal}{arXiv preprint arXiv:1807.07984}
  (\bibinfo{year}{2018}).
\bibitem[{{Wu} et~al.(2015){Wu}, {Song}, {Khosla}, {Yu}, {Zhang}, {Tang}, and
  {Xiao}}]{wu20153d}
\bibinfo{author}{Z.~{Wu}}, \bibinfo{author}{S.~{Song}},
  \bibinfo{author}{A.~{Khosla}}, \bibinfo{author}{F.~{Yu}},
  \bibinfo{author}{L.~{Zhang}}, \bibinfo{author}{X.~{Tang}},
  \bibinfo{author}{J.~{Xiao}},
\newblock \bibinfo{title}{3d shapenets: A deep representation for volumetric
  shapes},
\newblock in: \bibinfo{booktitle}{2015 IEEE Conference on Computer Vision and
  Pattern Recognition (CVPR)}, \bibinfo{year}{2015}, pp.
  \bibinfo{pages}{1912--1920}.

\end{thebibliography}

\end{document}